\documentclass[conference]{IEEEtran}
\IEEEoverridecommandlockouts

\usepackage{cite}
\usepackage{amsmath,amssymb,amsfonts}
\usepackage{algorithmic}

\usepackage{graphicx}
\usepackage{textcomp}
\usepackage{xcolor}
\usepackage{comment}
\usepackage{multirow}

\usepackage{algorithm}
\usepackage{subcaption}
\usepackage[bookmarks=false]{hyperref}
\usepackage[top=0.71in,bottom=1.0in,left=0.6251in,right=0.6251in]{geometry}
\def\BibTeX{{\rm B\kern-.05em{\sc i\kern-.025em b}\kern-.08em
    T\kern-.1667em\lower.7ex\hbox{E}\kern-.125emX}} 

\begin{document}

\bstctlcite{IEEEexample:BSTcontrol}

\title{SALT: A Lightweight Model Adaptation Method\\ for Closed Split Computing Environments\\
\thanks{This work was supported in part by JST-ALCA-Next Japan Grant Number JPMJAN24F1 and  JST PRESTO Grant Number JPMJPR2035.}
}

\author{
\IEEEauthorblockN{Yuya Okada, Takayuki Nishio}
\IEEEauthorblockA{\textit{Institute of Science Tokyo}, Tokyo, Japan}
\IEEEauthorblockA{okada.y.0249@m.isct.ac.jp, nishio@ict.e.titech.ac.jp}
}

\maketitle

\begin{abstract}
We propose SALT (Split-Adaptive Lightweight Tuning), a lightweight model adaptation framework for Split Computing under closed constraints, where the head and tail networks are proprietary and inaccessible to users.
In such closed environments, conventional adaptation methods are infeasible since they require access to model parameters or architectures.
SALT addresses this challenge by introducing a compact, trainable adapter on the client side to refine latent features from the head network, enabling user-specific adaptation without modifying the original models or increasing communication overhead.
We evaluate SALT on user-specific classification tasks with CIFAR-10 and CIFAR-100, demonstrating improved accuracy with lower training latency compared to fine-tuning methods.
Furthermore, SALT facilitates model adaptation for robust inference over lossy networks, a common challenge in edge–cloud environments.
With minimal deployment overhead, SALT offers a practical solution for personalized inference in edge AI systems under strict system constraints.
\end{abstract}

\begin{IEEEkeywords}
Split Computing, Edge AI, Lightweight Fine-tuning, Personalized Inference, Robustness to Packet Loss.
\end{IEEEkeywords}

\section{Introduction}
\label{sec:Introduction}
The increasing scale of deep learning models deployed in cloud-based AI services has raised concerns regarding server-side computational load and inference latency.
To address these challenges, Split Computing has emerged as a promising paradigm that offloads part of a large cloud-based model to the client device~\cite{kang, shao}.
In this architecture, the neural network model is partitioned into a head network executed on the client and a tail network retained on the cloud.
Input data is partially processed on the device, and the latent feature is transmitted to the cloud, where the final inference is performed.
This design alleviates cloud-side computation and reduces latency, while maintaining compatibility with the original model architecture and processing flow.

A key challenge in making Split Computing practical is the need to adapt the model specifically for Split Computing environments. Model adaptation can be broadly categorized into two aspects. The first is \textbf{model architecture adaptation}. Since many deep learning models map inputs to high-dimensional features, naively splitting a pre-trained model can lead to excessive communication overhead. Additionally, client devices often have limited computational resources, making it difficult to process large head networks. To address these issues, prior work has explored optimizing the head network architecture to reduce communication and computational costs~\cite{shimizu, chen}. The second is \textbf{model parameter adaptation}, which tunes the parameters of the given models—often the head network—to the user’s data distribution, task, or network conditions, improving inference accuracy and communication efficiency without altering the model structure. In particular, pre-tuning the model for lossy networks has been shown to enhance loss robustness, enabling accurate inference despite packet loss while reducing retransmission overhead~\cite{itahara}. This paper focuses on the latter, model parameter adaptation.

Model parameter adaptation—hereafter referred to simply as model adaptation for brevity—faces the key challenge of tuning model parameters efficiently in terms of both computation and communication, and various approaches have been proposed to address this. TinyTL enables efficient on-device fine-tuning by freezing large weights and updating only small bias modules, allowing lightweight adaptation to personalized data~\cite{cai}. PockEngine leverages sparse backpropagation and graph optimization to reduce the training cost on edge devices~\cite{zhu}. Additionally, although this approach also involves model architecture adaptation, a pre-trained head network can be used to train a new head network via knowledge distillation, reducing both computational and communication burdens while adapting to the target domain~\cite{matsubara2}.

While these studies have demonstrated promising results for model adaptation, they typically assume access to the internal structure or parameters of the given model.
In real-world settings, however, models are increasingly deployed in a closed form—distributed as pre-trained modules or accessed only via restricted APIs—to protect intellectual property and prevent malicious exploitation. Given the growing adoption of AI technologies in commercial and privacy-sensitive applications, this trend is expected to continue. As a result, there is an increasing demand for adaptation methods that can function effectively under such closed model constraints.

To overcome this limitation, we propose SALT (Split-Adaptive Lightweight Tuning), a novel framework for enabling model adaptation in Split Computing environments under closed constraints.
The core idea of SALT is inspired by recent work in computer vision and large language models ~\cite{houlsby,rebuffi} and ~\cite{hu}, which enables low-cost model adaptation by inserting small trainable modules into otherwise frozen networks.
SALT extends this concept to the closed-model setting of Split Computing. Specifically, SALT introduces a compact, trainable adapter on the client side to refine latent features from the head network, enabling model adaptation without modifying the original models or increasing communication overhead.
Through extensive experiments, we demonstrate that SALT achieves high accuracy and significantly reduces training latency compared to existing approaches, while maintaining stable performance even under severe packet loss.

\section{Lightweight Model Adaptation for Split Computing}
\label{sec:Lightweight Model Adaptation for Split Computing}
\subsection{System Model}

We consider a user-specific inference scenario in edge AI applications, where resource-constrained client devices such as smartphones or embedded edge devices perform personalized image recognition tasks.
Input data $x$ remains on the client side, while the corresponding labels $y$ are stored on the cloud server, reflecting privacy-preserving data processing requirements.

We assume a Split Computing system in which a pre-trained deep neural network is divided into a head network $H(\cdot)$ executed on the client and a tail network $T(\cdot)$ on the server. The client and server communicate over a network with a bandwidth of $80~\mathrm{Mbit/s}$, reflecting practical edge AI deployment conditions. In this architecture, the client processes an input sample $x$ using the head network to produce a latent feature $z = H(x)$, which is transmitted to the server. The server computes the final prediction $y = T(z)$ using the tail network.

The following assumptions are also made in this study:

\noindent\textbf{Model assumption:}
The architectures of $H(\cdot)$ and $T(\cdot)$ are pre-designed and pre-optimized for Split Computing. The head network is lightweight enough to run on the client device, and appropriate bottleneck structures are inserted at the split point to reduce the dimensionality of $z$ for efficient transmission. The internal parameters and structures of $H(\cdot)$ and $T(\cdot)$ cannot be accessed or modified, reflecting scenarios where the models are provided as proprietary modules or accessed via APIs.

\vspace{1mm}
\noindent\textbf{Data assumption:}
Each client holds only a small amount of unlabeled, user-specific data, reflecting limited local data collection and privacy-preserving practices. Ground-truth labels corresponding to this data are assumed to reside on the server side, a setting commonly adopted in split learning and related collaborative learning frameworks~\cite{gupta}. We assume that user-specific data represents a narrow, user-defined domain of interest that may differ from the data distribution used to train the original model. The goal is to adapt the model to this local data without requiring data sharing or centralized aggregation.

\vspace{1mm}
\noindent\textbf{Communication assumption:}
Communication between the client and server may be subject to packet loss. Following prior work~\cite{itahara}, we model packet loss as element-wise random drops in the transmitted latent feature; the details are described in Section~\ref{sec:training}. Consequently, improving robustness to packet loss—enabling inference without retransmission under lossy conditions—becomes one of the goals of model adaptation.

\subsection{Overview of SALT in Split Computing}
\begin{figure}[t]
    \hspace{-3mm}
    \centerline{\includegraphics[width=0.9\linewidth]{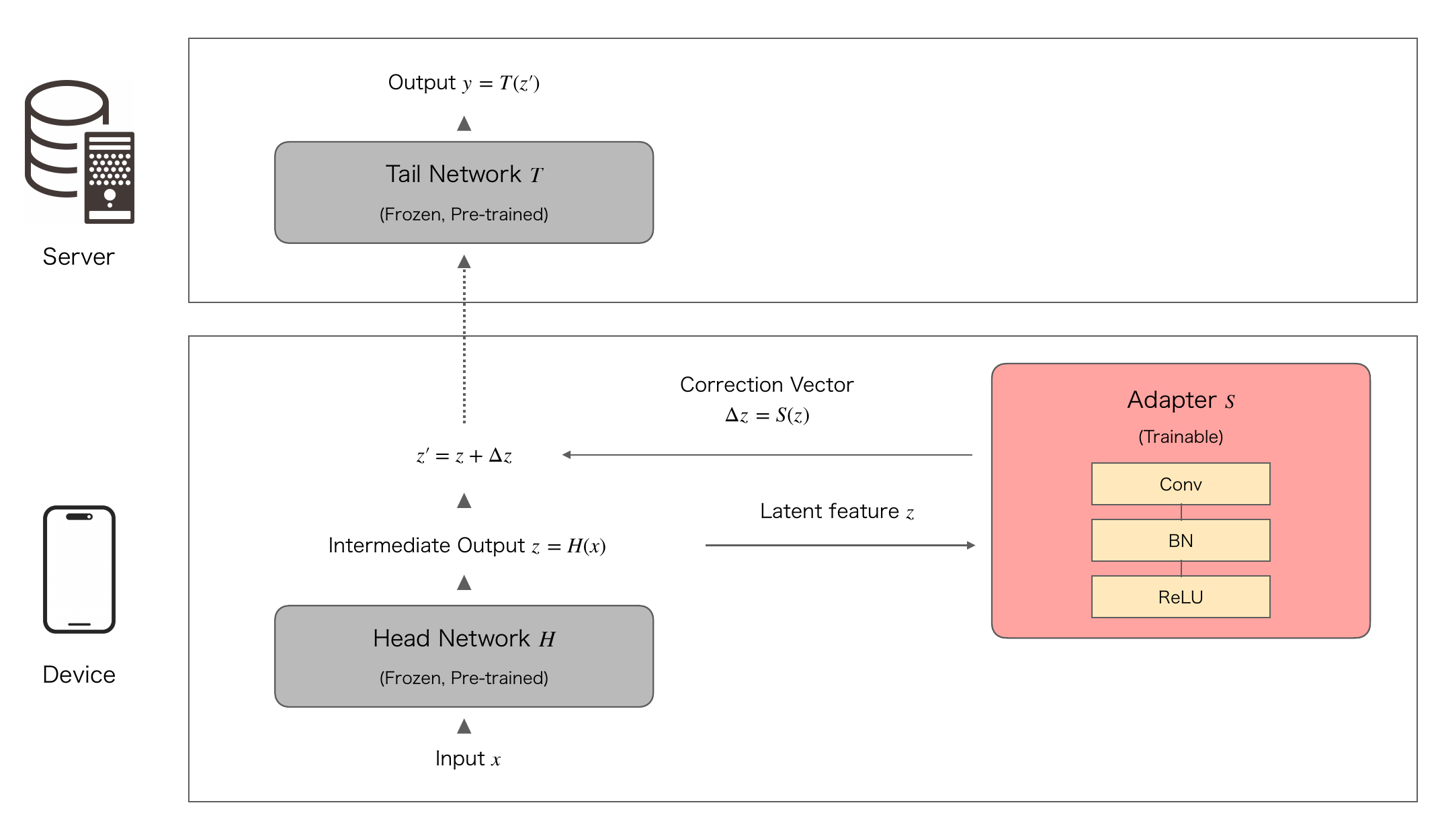}}
    \caption{Architecture of Split Computing and SALT with Residual Adapter}
    \label{fig:architecture}
    \vspace{-4.138mm}
\end{figure}

Our goal is to realize a model adaptation method for Split Computing that can be applied even in closed scenarios, where the structures and parameters of the head and tail networks are proprietary and inaccessible to users. To achieve this, we propose Split-Adaptive Lightweight Tuning (SALT). Figure~\ref{fig:architecture} illustrates the architecture of Split Computing with SALT. 
SALT introduces a lightweight trainable module, referred to as an Adapter $S(\cdot)$, between the client-side closed head network and the server-side closed tail network, enabling adaptation to user-specific data distributions and communication losses without modifying the existing models.
The Adapter estimates a correction vector $\Delta z = S(z)$ from the latent feature $z$ produced by the head network $H(x) = z$. 
This correction vector is then added to the original feature to obtain a refined representation $z' = z + \Delta z$, 
which is used to enhance user-specific inference accuracy.

To support various adaptation scenarios, SALT offers two architectural options as follows:\\
\noindent\textbf{Residual Adapter (Fig.~\ref{fig:architecture}):}  
This adapter operates on the latent feature $z = H(x)$ by estimating a correction vector $\Delta z = S(z)$ and applying it to obtain the refined feature $z' = z + \Delta z$.  
Due to the residual formulation, this structure promotes faster convergence and improved training stability.  
In this paper, we refer to this Residual Adapter simply as the ``Adapter'', and adopt it as the default design for its effectiveness in both computational efficiency and training performance.

\vspace{1mm}
\noindent\textbf{Insertion Adapter:} 
This variant appends trainable convolutional layers directly after the latent feature $z$ output by the head network on the client side, to estimate the corrected feature $z'$. 
This structure is equivalent to augmenting the head network with additional layers.
The number of layers and trainable parameters are kept comparable to the Residual Adapter-type.
Some existing studies enhance the head network by adding layers such as bottleneck structures to reduce communication overhead.
However, this approach requires a decoder on the server side to restore the original feature dimension before inputting it into the tail network.
Since this would violate our assumption that the tail network is shared and fixed across clients, we do not adopt bottleneck structures in this work.

\subsection{Training Procedure for SALT}\label{sec:training}
In this section, we describe the training procedure for Adapter. The head and tail networks are treated as fixed, pre-trained closed models that remain unchanged during training. Only the parameters of Adapter are updated, with the dual objectives of improving inference accuracy on user-specific data distributions and enhancing robustness to feature degradation caused by transmission errors.

The training process is outlined in Algorithm~\ref{alg:salt_training}. In this setting, the client is responsible for generating latent features and updating Adapter, while the server computes the loss and returns the gradients required for training. Following standard practices in split learning~\cite{gupta}, training is performed in mini-batches; for clarity, we describe the procedure on a per-sample basis. Specifically, the client computes the intermediate latent feature $z$ by applying the frozen head network $H(\cdot)$ to the input $x$. Adapter $S(\cdot)$ then produces a correction vector $\Delta z$, which is added to $z$ to obtain the adjusted latent feature $z' = z + \Delta z$. This adjusted feature is transmitted over the client-to-server communication channel, which may introduce degradation modeled by $C_{\mathrm{c \rightarrow s}}(\cdot)$, yielding the received feature $\tilde{z}'$.

Following prior work~\cite{itahara}, $C_{\mathrm{c \rightarrow s}}(\cdot)$ models communication degradation due to packet loss. We assume that latent features are transmitted in interleaved packets to mitigate burst loss; as a result, packet loss manifests as random missing entries in the latent feature vector. During forward propagation, these missing values are zero-filled to allow processing to continue. Formally, $C_{\mathrm{c \rightarrow s}}(\cdot)$ is represented by a probabilistic mask function $m(p)$, where $p$ is the packet loss rate. The degraded latent feature after transmission is thus given by $\tilde{z}' = z' \odot m(p)$, where $\odot$ denotes element-wise (Hadamard) product. By training under this degraded channel, the SALT module learns to compensate for communication errors, thereby improving inference robustness against packet loss.

Upon receiving $\tilde{z}'$, the server computes the final prediction $\hat{y}$ using the frozen tail network $T(\cdot)$. The prediction is compared to the ground-truth label $y$ to calculate the loss $\mathcal{L}$. Since the server only has access to $T(\cdot)$, it can compute gradients only up to $\partial \mathcal{L} / \partial \tilde{z}'$, denoted as $\delta$. This gradient is transmitted back to the client via the server-to-client communication channel, potentially subject to degradation modeled by $C_{\mathrm{s \rightarrow c}}(\cdot)$, yielding the received gradient $\tilde{\delta}$. In this work, $C_{\mathrm{s \rightarrow c}}(\cdot)$ is assumed to be an identity function, as retransmission control ensures reliable delivery to stabilize training. Investigating lossy server-to-client channels is left as future work.

Finally, the client continues backpropagation from $\tilde{\delta}$ through $S(\cdot)$ to update Adapter’s parameters. This training approach enables Adapter to be adapted to both the user’s data distribution and the communication channel, achieving model adaptation without accessing or modifying the frozen head and tail networks, and thus preserving model confidentiality.

\begin{algorithm}[t]
\caption{SALT training procedure based on split learning}
\label{alg:salt_training}
\begin{algorithmic}[1]
\REQUIRE Frozen head network $H(\cdot)$, Frozen tail network $T(\cdot)$, Trainable SALT module $S(\cdot)$, Training dataset $\mathcal{D} = \{(x_k, y_k)\}$
\FOR{each epoch}
    \FOR{each mini-batch $B = \{(x_k, y_k)\}$ in $\mathcal{D}$}
        \STATE \textbf{Client side:}
        \STATE \hspace{1.5em} $Z \leftarrow \{ H(x_k) \mid x_k \in B \}$
        \STATE \hspace{1.5em} $\Delta Z \leftarrow \{ S(z_k) \mid z_k \in Z \}$
        \STATE \hspace{1.5em} $Z' \leftarrow \{ z_k + \Delta z_k \mid z_k \in Z, \Delta z_k \in \Delta Z \}$
        \STATE \hspace{1.5em} Transmit $\tilde{Z}'$ to server
        \STATE \textbf{Server side:}
        \STATE \hspace{1.5em} Receive $\tilde{Z}'= C_{\mathrm{c \rightarrow s}}(Z')$
        \STATE \hspace{1.5em} $\hat{Y} \leftarrow \{ T(\tilde{z}'_k) \mid \tilde{z}'_k \in \tilde{Z}' \}$
        \STATE \hspace{1.5em} $\mathcal{L} \leftarrow \frac{1}{|B|} \sum_{x_k \in B} \text{CrossEntropy}(\hat{y}_k, y_k)$
        \STATE \hspace{1.5em} $\delta \leftarrow$ gradients of $\mathcal{L}$ w.r.t. $\tilde{Z}'$
        \STATE \hspace{1.5em} Transmit $\delta$ to client
        \STATE \textbf{Client side:}
        \STATE \hspace{1.5em} Receive $\tilde{\delta}= C_{\mathrm{s \rightarrow c}}(\delta)$
        \STATE \hspace{1.5em} Update parameters of $S(\cdot)$ using $\tilde{\delta}$
    \ENDFOR
\ENDFOR
\end{algorithmic}
\end{algorithm}

\section{Performance Evaluation}
\label{sec:Performance Evaluation}
\subsection{Setup}
\noindent\textbf{Datasets and Tasks:}
The broader goal of SALT is to handle domain shifts at the input level, such as those caused by grayscale or infrared imaging, or harsh lighting conditions on edge devices. In this study, however, we focus on feature-level adaptation and evaluate SALT using two standard image classification benchmarks: CIFAR-10 and CIFAR-100~\cite{cifar}. 

To construct the pre-trained head and tail networks, we first trained a standard ResNet-18 model on the full CIFAR-10 and CIFAR-100 datasets.
For CIFAR-10, the model was trained on all 10 classes using $50,000$ training images and evaluated on a 10-class classification task with $10,000$ test images, achieving a top-1 accuracy of $89.4\%$.
For CIFAR-100, models were trained on all 100 classes using the same number of training and test samples and tested on a 100-class classification task, achieving a top-1 accuracy of $65.7\%$.
These pre-trained models serve as the basis for all experiments.
In our proposed method (SALT), which assumes closed model constraints, the head and tail networks are fixed, reflecting realistic edge–cloud environments where commercial models are deployed as closed components.
In contrast, baselines such as Retrain and Tune are allowed to modify or fine-tune the head network to simulate access to open model components.

\begin{figure}[t]
\centering
    \includegraphics[width=0.9\linewidth]{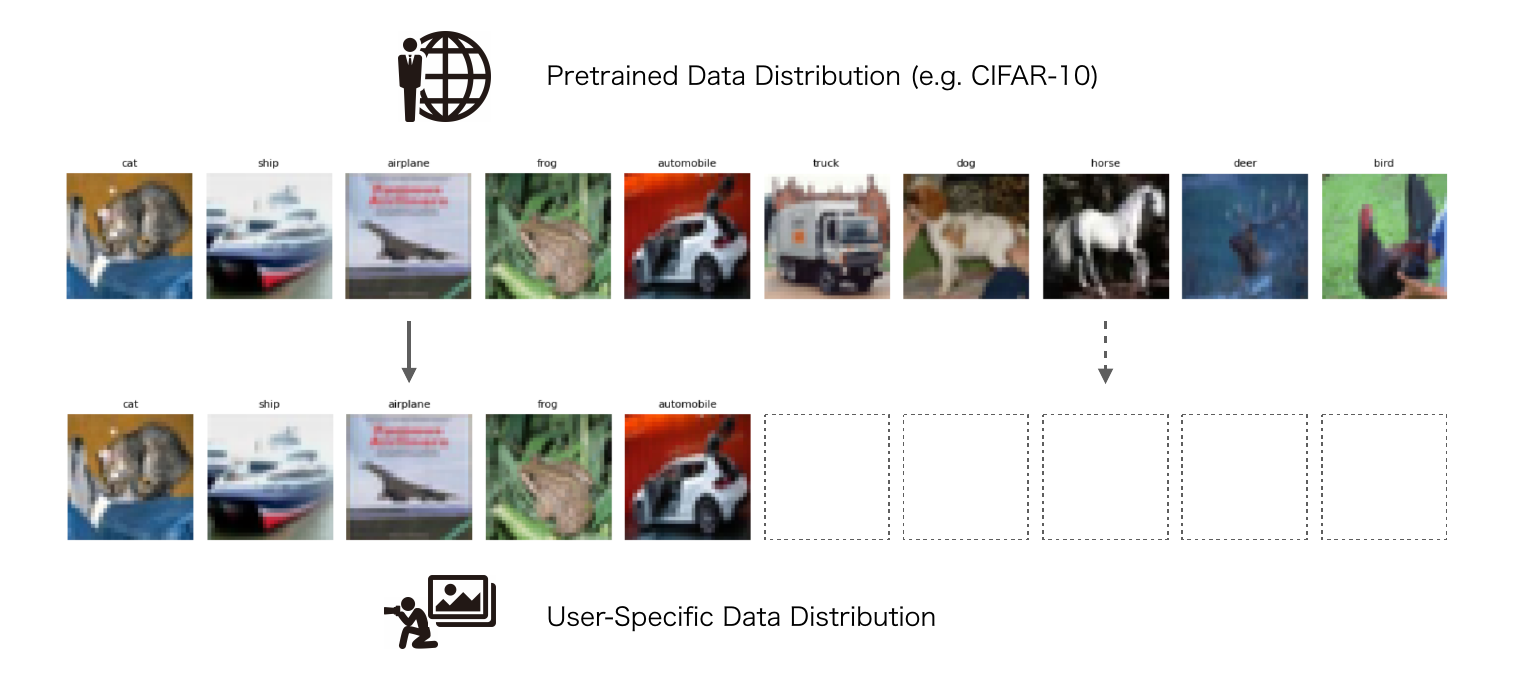}
    \caption{Illustration of the adaptation setting. While the model is pre-trained on a complete dataset (e.g., CIFAR-10 with 10 classes), only a subset of these classes is used to simulate user-specific data.}
    \label{fig:adaptation_setting}
    \vspace{-2mm}
\end{figure}

To simulate user-specific adaptation, we construct sub-tasks by selecting a subset of classes from each dataset. 
For CIFAR-10, we use five classes [airplane, automobile, bird, cat, deer] (labels 0–4), resulting in $25,000$ training and $5,000$ test images, as illustrated in Figure~\ref{fig:adaptation_setting}. 
For CIFAR-100, we select ten classes [apple, aquarium fish, baby, bear, beaver, bed, bee, beetle, bicycle, bottle] (labels 0–9), using $5,000$ training and $1,000$ test images.
This setup simulates adapting a globally pre-trained model to a user-specific domain of interest.

Using these sub-tasks, we evaluate our proposed method (SALT) and several baselines in a setting where a globally pre-trained model is locally adapted to user-specific data.
All methods are trained or fine-tuned on the corresponding sub-task training images and evaluated on the corresponding test images. 

\begin{figure}[t]
    \centerline{\includegraphics[width=0.95\linewidth]{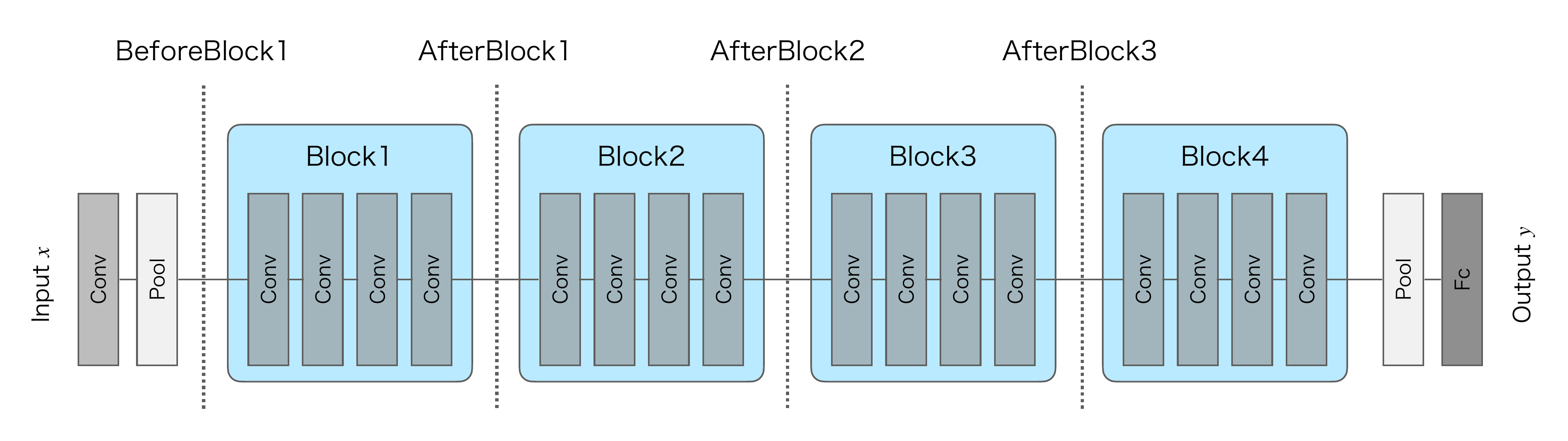}}
    \caption{Definition of split points in ResNet-18}
    \label{fig:splitpoints}
    \vspace{-3.3mm}
\end{figure}

\vspace{1mm}
\noindent\textbf{Model:}
ResNet-18~\cite{resnet} was used as the backbone network and was split into head and tail networks at one of the four split points shown in Fig.~\ref{fig:splitpoints}.
The choice of split point affects both the client-side computation cost and the communication overhead.

Unless otherwise specified, all experiments are conducted with the split point fixed at AfterBlock2 in the ResNet-18 architecture.

Adapter is implemented as a lightweight three-layer convolutional network.
Specifically, it first applies a $3\times3$ convolutional layer, followed by batch normalization and a ReLU activation function, without changing the number of channels.
This sequence is repeated twice, and then a $1\times1$ convolutional layer is applied to produce the final correction output $\Delta z$.
Throughout the Adapter, the number of channels is kept the same as that of the input latent feature; for example, $128$ channels are maintained at the split point AfterBlock2.

Notably, Adapter introduces only a modest increase in model complexity compared to the head network. In our implementation, Adapter contains approximately $443\mathrm{K}$ parameters and requires $28.39$ million multiply-accumulate operations (MMACs) per forward pass. In contrast, the head network—defined as the ResNet-18 backbone up to a selected split point—has between $1.73\mathrm{K}$ and $2.78\mathrm{M}$ parameters, depending on the split configuration. For example, at the deepest split point used in our experiments (AfterBlock3), the head network consists of conv1 through layer3 and contains approximately $2.78\mathrm{M}$ parameters and $54.72\mathrm{MMACs}$.

\vspace{1mm}
\noindent\textbf{Hyperparameters:}
We employed the Adam optimizer and used cross-entropy loss computed on the output of the frozen tail network. Training was conducted with a batch size of $128$ and a learning rate of $1\mathrm{e-}3$, and continued until convergence, defined as the point where the validation loss stopped improving. All models were trained for up to $100$ epochs with early stopping. Input images were standardized to zero mean and unit variance. We used ReLU as the activation function and applied max pooling. For all experiments, training was performed on an NVIDIA Quadro RTX 6000 GPU. The split points considered in the ResNet-18 architecture were BeforeBlock1, AfterBlock1, AfterBlock2, and AfterBlock3. To evaluate robustness under communication degradation, we simulated packet loss rates of $0.0$, $0.25$, $0.5$ and $0.75$ over a $80~\mathrm{Mbit/s}$ bandwidth network. A summary of the training configuration is provided in Table~\ref{tab:training_config}.

\begin{table}[t]
    \centering
    \caption{Training configuration and environment used for all experiments.}
    \label{tab:training_config}
    \resizebox{0.925\linewidth}{!}{%
    \begin{tabular}{|c |c |}
    \hline
    \textbf{Configuration Item} & \textbf{Details} \\
    \hline
    Optimizer          & Adam \\
    Loss Function      & Cross Entropy Loss (on tail network outputs) \\
    Learning Rate      & $1\mathrm{e-}3$ \\
    Batch Size         & $128$ \\
    Training Epochs    & $100$ (or until validation loss plateaus) \\
    Activation Function & ReLU \\
    Pooling Type       & Max pooling \\
    Standardization    & Input standardized (mean 0, std 1) \\
    Split Points       & BeforeBlock1, AfterBlock1, AfterBlock2, AfterBlock3 \\
    Packet Loss Rates         & $0.0, 0.25, 0.5, 0.75$ (simulated packet drop scenarios) \\
    Network Bandwidth & $80~\mathrm{Mbit/s}$ \\
    GPU Environment    & NVIDIA Quadro RTX 6000 \\
    \hline
    \end{tabular}
    }
    \vspace{-2mm}
\end{table}

\vspace{1mm}
\noindent\textbf{Communication Model:}
To assess the communication cost in our edge–cloud setup, we adopt a simplified model of data transmission between the client and the server.
The communication latency is estimated numerically by calculating the transmission latency based on the size of the latent feature, the batch size, and the available network bandwidth.
However, propagation delay is not considered as an error.

We assume that latent feature is transmitted during both forward and backward passes in Split Computing, and compute the per-batch communication latency accordingly as follows:
\begin{equation}
T_{\text{comm/batch}} = 2 \times \left( \frac{V_z \cdot B}{\theta} \right),
\label{eq:comm_round}
\end{equation}
where $V_z$ is the latent feature size in bits, $B$ is the batch size, and $\theta$ denotes the network bandwidth.
Based on the configuration shown in Table~\ref{tab:training_config}.

The total communication latency across training is then calculated as:
\begin{equation}
T_{\text{comm-total}} = E \times \left\lceil \frac{N}{B} \right\rceil \times T_{\text{comm/batch}},
\label{eq:comm_total}
\end{equation}
where $E$ is the number of epochs determined by early stopping, and $N$ is the number of training samples.
This formulation~\eqref{eq:comm_total} enables us to quantify the impact of communication overhead on training efficiency under realistic edge AI constraints.

In the simulation, packet loss is modeled using the same probabilistic mask $m(p)$ described in the training procedure (Section~\ref{sec:training}), where each element of the transmitted latent feature is independently dropped and subsequently zero-filled with probability $p$. Unless otherwise specified for experiments explicitly evaluating robustness under communication degradation, we set $p = 0$.

\vspace{1mm}
\noindent\textbf{Baselines:}
To evaluate the effectiveness of our proposed method (SALT), we compare it against two types of methods: (1) common adaptation baselines that assume varying levels of model access, and (2) architectural variants of SALT designed to test different integration styles. 
The common baselines include Original (no adaptation), Retrain (training the head from scratch), and Tune (fine-tuning the pre-trained head). 
In addition, we include two variants of our method—SALT (Insertion Adapter) and SALT (Residual Adapter)—to investigate the impact of Adapter placement and structure. 
Table~\ref{tab:baselines} summarizes all compared methods along with their access assumptions.

\begin{table}[t]
  \centering
  \small
  \caption{Comparison of baseline methods}
  \label{tab:baselines}
  \resizebox{0.95\linewidth}{!}{%
  \begin{tabular}{|c |c |c |c |}
    \hline
    Method & Description & Head Access & Tail Access \\
    \hline
    Original & Without adaptation & $\times$ & $\times$ \\
    Retrain & Train head from scratch & $\bigcirc$ (train) & $\times$ \\
    Tune & Fine-tune pre-trained head & $\bigcirc$ (fine-tune) & $\times$ \\
    SALT (Insertion Adapter) & Insert Adapter & $\times$ & $\times$ \\    
    SALT (Residual Adapter) & Residual Adapter & $\times$ & $\times$ \\
    \hline
  \end{tabular}
  }
  \vspace{-2mm}
\end{table}

\vspace{1mm}
\noindent\textbf{Metrics:}
We evaluate model performance in terms of classification accuracy on user-specific tasks, total training latency, and inference latency. 

For accuracy, we report the Personalized Top-1 Test Accuracy, measured on the subset of classes assigned to each user: five classes for CIFAR-10 and ten classes for CIFAR-100.
This metric reflects the model’s adaptation performance to the user-specific class subset, rather than accuracy on the full dataset.
Total training latency is defined as the total time required for model convergence, computed as the sum of the measured computation latency and the estimated communication latency, assuming a network bandwidth of $80~\mathrm{Mbit/s}$.
Inference latency is measured as the time required to perform inference on a single batch of $128$ samples.

All results are averaged over $10$ independent trials under various split point and packet loss conditions.

\subsection{Results and Discussion}
\subsubsection{Accuracy vs. Total Training Latency}
\begin{figure}[t]
  \centering
  \begin{minipage}{\linewidth}
    \centering
    \hspace{-5mm}
    \includegraphics[width=0.85\linewidth, keepaspectratio]{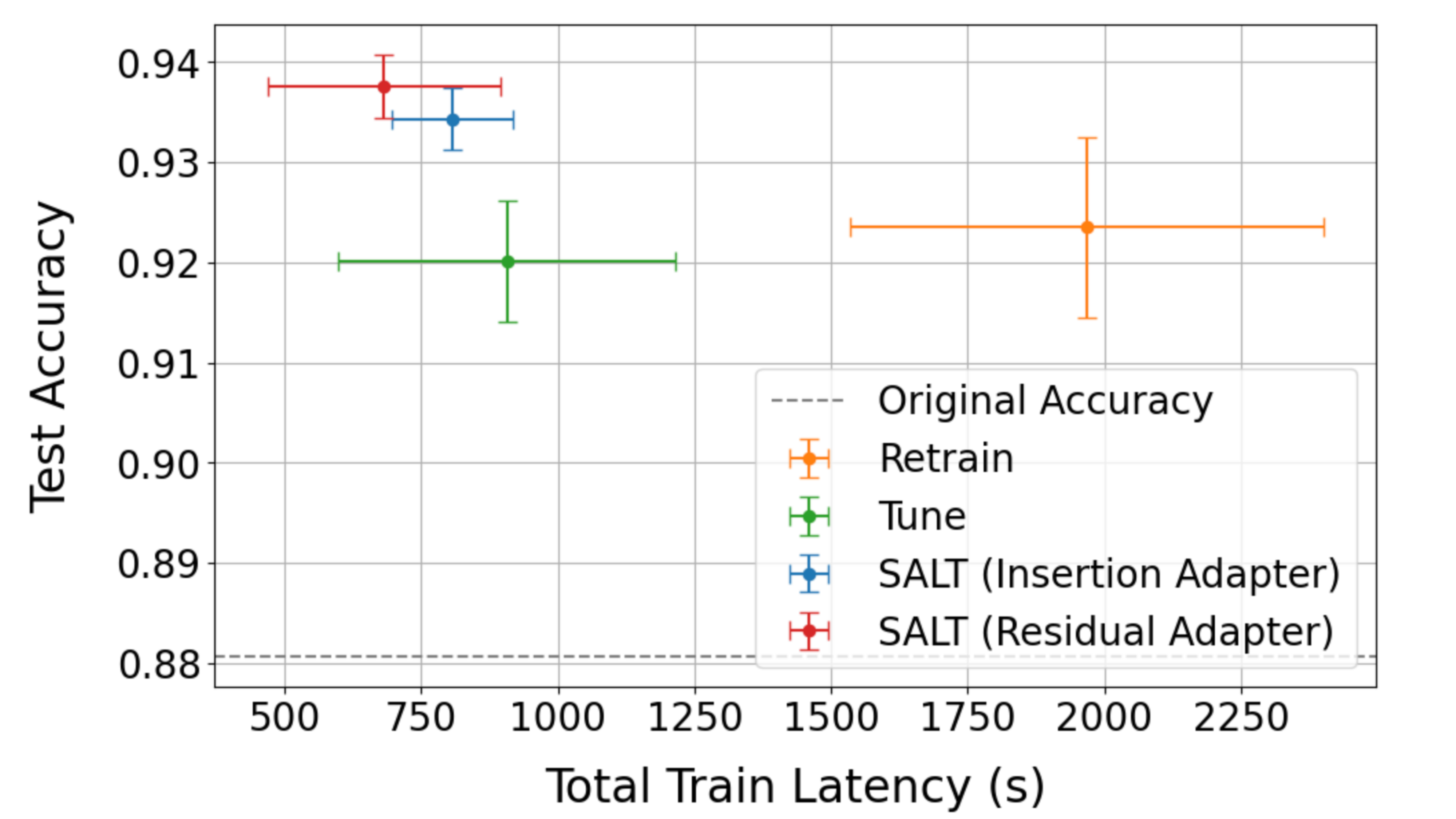}
    \subcaption{Results on CIFAR-10}
    \vspace{2mm}
  \end{minipage}

  \begin{minipage}{\linewidth}
    \centering
    \hspace{-5mm}
    \includegraphics[width=0.853\linewidth, keepaspectratio]{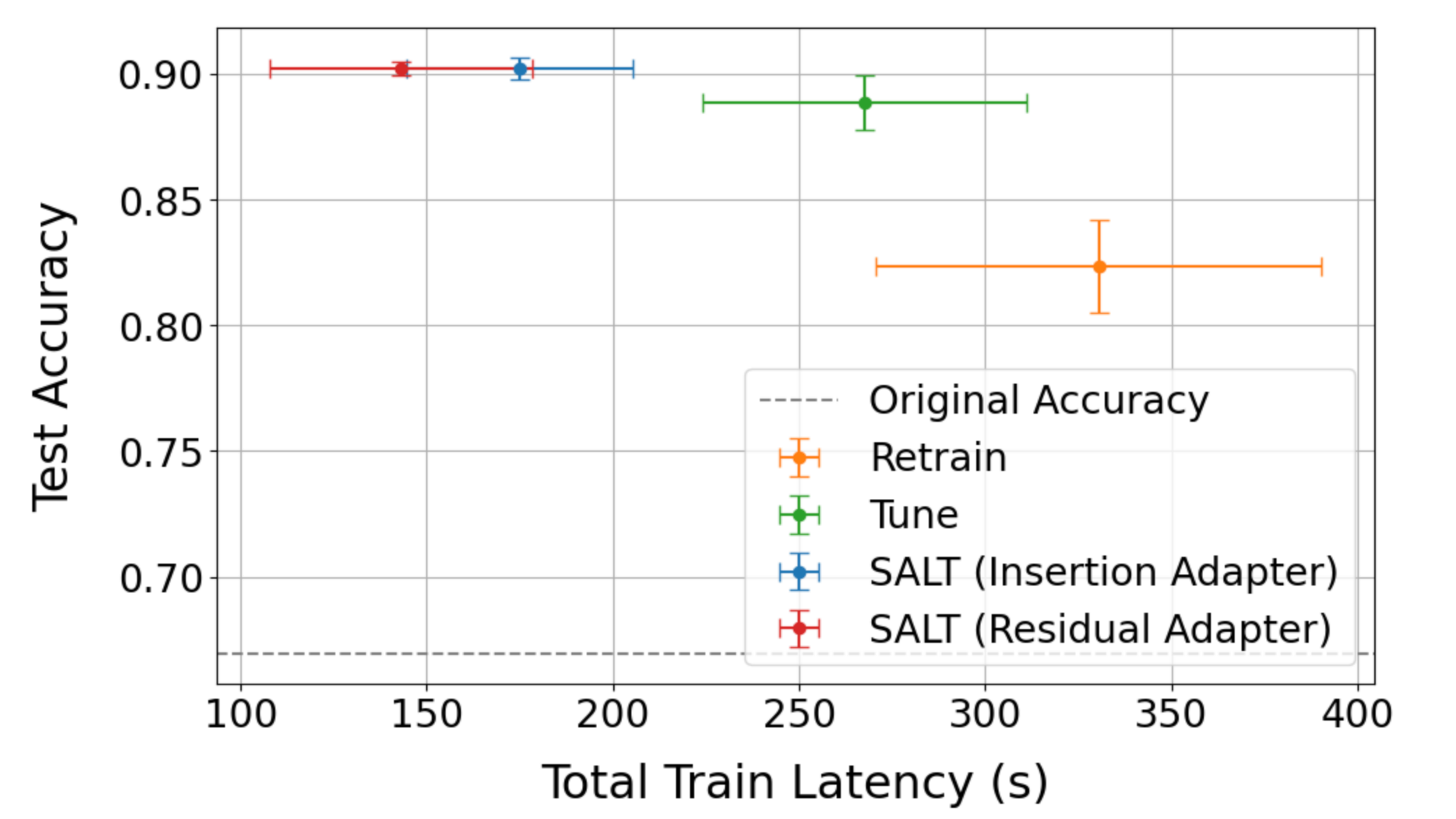}
    \subcaption{Results on CIFAR-100}
  \end{minipage}

  \caption{Accuracy vs. Total Training Latency for each method. SALT maintains high accuracy with low training cost. Error bars indicate $95\%$ confidence intervals over 10 trials.}
  \label{fig:cifar10_100_scatter}
  \vspace{-2mm}
\end{figure}

Table~\ref{tab:total_results} and Figure~\ref{fig:cifar10_100_scatter} present the performance comparison of various methods on CIFAR-10 and CIFAR-100.  
Table~\ref{tab:total_results} summarizes the accuracy, number of training epochs, computation latency(Comp. Latency), communication latency(Comm. Latency), and inference latency(Inf. Latency), with the computation and communication components shown separately.  
Figure~\ref{fig:cifar10_100_scatter} visualizes the trade-off between accuracy and total training latency.  

These results demonstrate that SALT (both Insertion Adapter and Residual Adapter types) achieves comparable or higher accuracy than Retrain and Tune, while requiring significantly less training effort.

As shown in Figure.~\ref{fig:cifar10_100_scatter}(a), SALT (Residual Adapter) achieved the highest accuracy of $93.8\%$, outperforming Retrain ($92.4\%$) and Tune ($92.0\%$).  
Additionally, it converged in just $20.5$ epochs, requiring $682.1~\mathrm{s}$ of total training latency, which is significantly less than Retrain ($1967.9~\mathrm{s}$) and Tune ($906.9~\mathrm{s}$).

Although SALT introduces a trainable module, it does not increase inference latency—$5.3~\mathrm{s}$ compared to the original model’s $5.3~\mathrm{s}$—indicating that the accuracy improvements are achieved without sacrificing runtime efficiency.

To further assess the effectiveness of SALT, we applied the same experimental setup to CIFAR-100, which has more classes and fewer samples per class compared to CIFAR-10.  
As shown in Figure~\ref{fig:cifar10_100_scatter}(b), SALT (Residual Adapter) achieved $90.2\%$ accuracy, outperforming Retrain ($82.4\%$) and Tune ($88.9\%$), and converged in only $21.2~\mathrm{epochs}$ with $143.0~\mathrm{s}$ of total training latency.  
These results suggest that SALT can provide additional benefits in tasks with limited user-specific data per class.

\begin{table*}[t]
  \centering
  \small
  \caption{Performance Comparison of Adaptation Methods on CIFAR-10 and CIFAR-100}
  \label{tab:total_results}
  \resizebox{0.95\linewidth}{!}{%
  \begin{tabular}{|c |cccc|cccc|}
    \hline
    \multirow{2}{*}{Method} & \multicolumn{4}{c|}{CIFAR-10} & \multicolumn{4}{c|}{CIFAR-100} \\
    \cline{2-9}
    & Accuracy & Epochs & Comp./Comm. Latency [s] & Inf. Latency [s] & Accuracy & Epochs & Comp./Comm. Latency [s] & Inf. Latency [s] \\
    \hline
    Original & $0.881$ & - & - / - & $5.3$ & $0.669$ & - & - / - & $1.0$ \\
    Retrain & $0.924$ & $54.0$ & $857.7$ / $1110.2$ & $5.4$ & $0.824$ & $49.6$ & $122.1$ / $208.3$ & $1.0$ \\
    Tune & $0.920$ & $26.0$ & $372.3$ / $534.6$ & $5.4$ & $0.889$ & $39.6$ & $101.3$ / $166.3$ & $1.0$ \\
    SALT (Insertion Adapter) & $0.934$ & $25.3$ & $287.6$ / $520.2$ & $5.3$ & $0.902$ & $27.2$ & $60.7$ / $114.2$ & $1.0$ \\
    SALT (Residual Adapter) & $0.938$ & $20.5$ & $260.6$ / $421.5$ & $5.3$ & $0.902$ & $21.2$ & $54.0$ / $89.0$ & $1.0$ \\
    \hline
  \end{tabular}
  }
  \vspace{-2mm}
\end{table*}

\subsubsection{Robustness to Packet Loss}
\begin{figure}[t]
    \hspace{-2mm}
    \centerline{\includegraphics[width=0.85\linewidth]{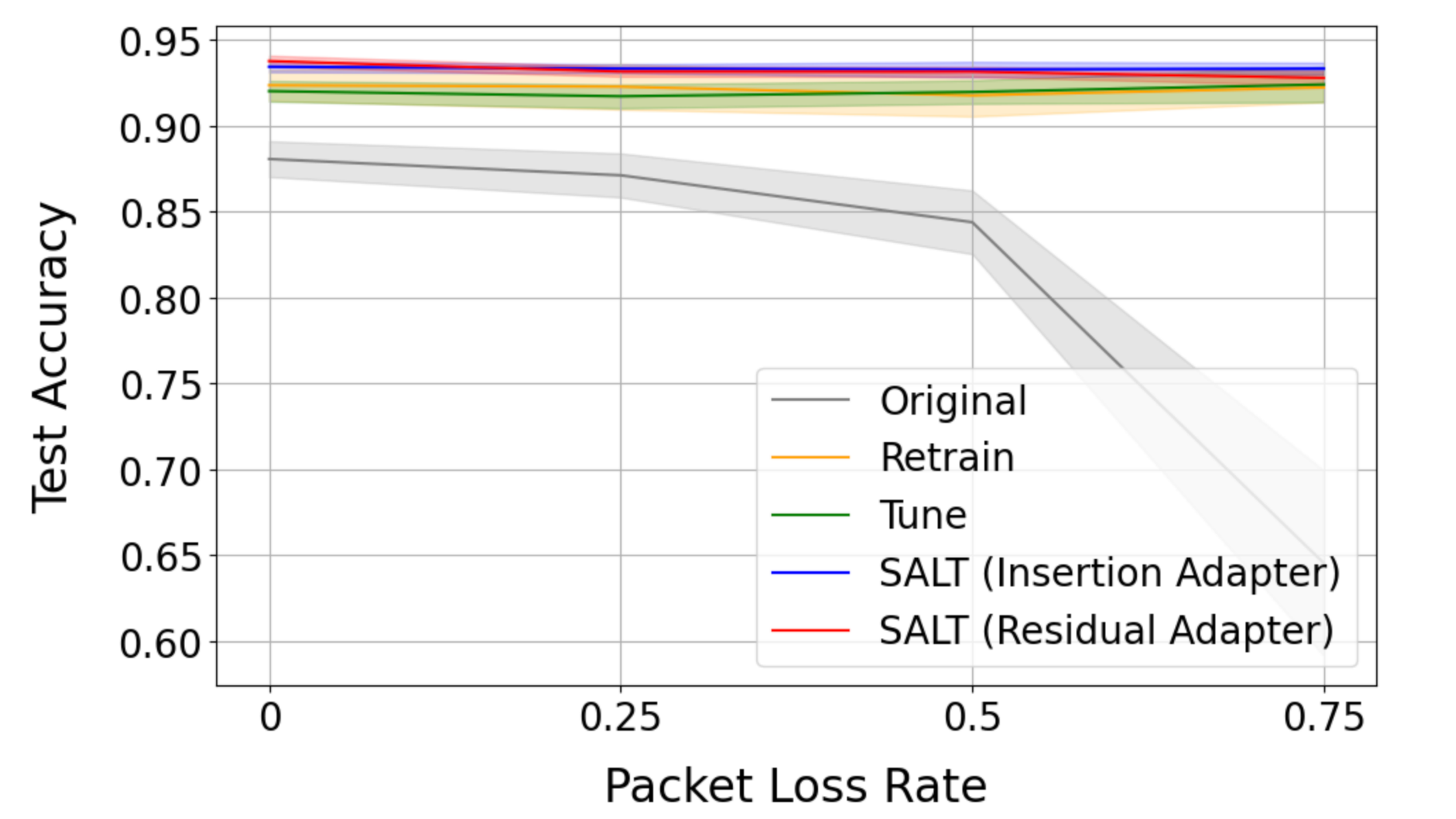}}
    \caption{Accuracy under varying packet loss rates for each method on CIFAR-10.}
    \label{fig:lossrate_accuracy}
    \vspace{-2mm}
\end{figure}

Next, we evaluate the improvement in packet loss robustness, which is another objective of model adaptation.  
Figure~\ref{fig:lossrate_accuracy} shows the personalized accuracy of each method as the packet loss rate increases from $0.0$ to $0.75$.
As expected, the Original model without any adaptation shows significant accuracy degradation as packet loss increases.
In contrast, Retrain, Tune, and both SALT variants exhibit a certain degree of robustness against packet loss, maintaining relatively high accuracy even under severe conditions.
Among these methods, SALT maintains competitive or superior accuracy compared to Retrain and Tune, while achieving much shorter total training latency.
These results highlight that adaptation with SALT not only enhances robustness to communication degradation but also enables efficient model adaptation with minimal training cost.

\subsubsection{Impact of Split Point on Accuracy}
\begin{figure}[t]
    \hspace{-2mm}
    \centerline{\includegraphics[width=0.85\linewidth]{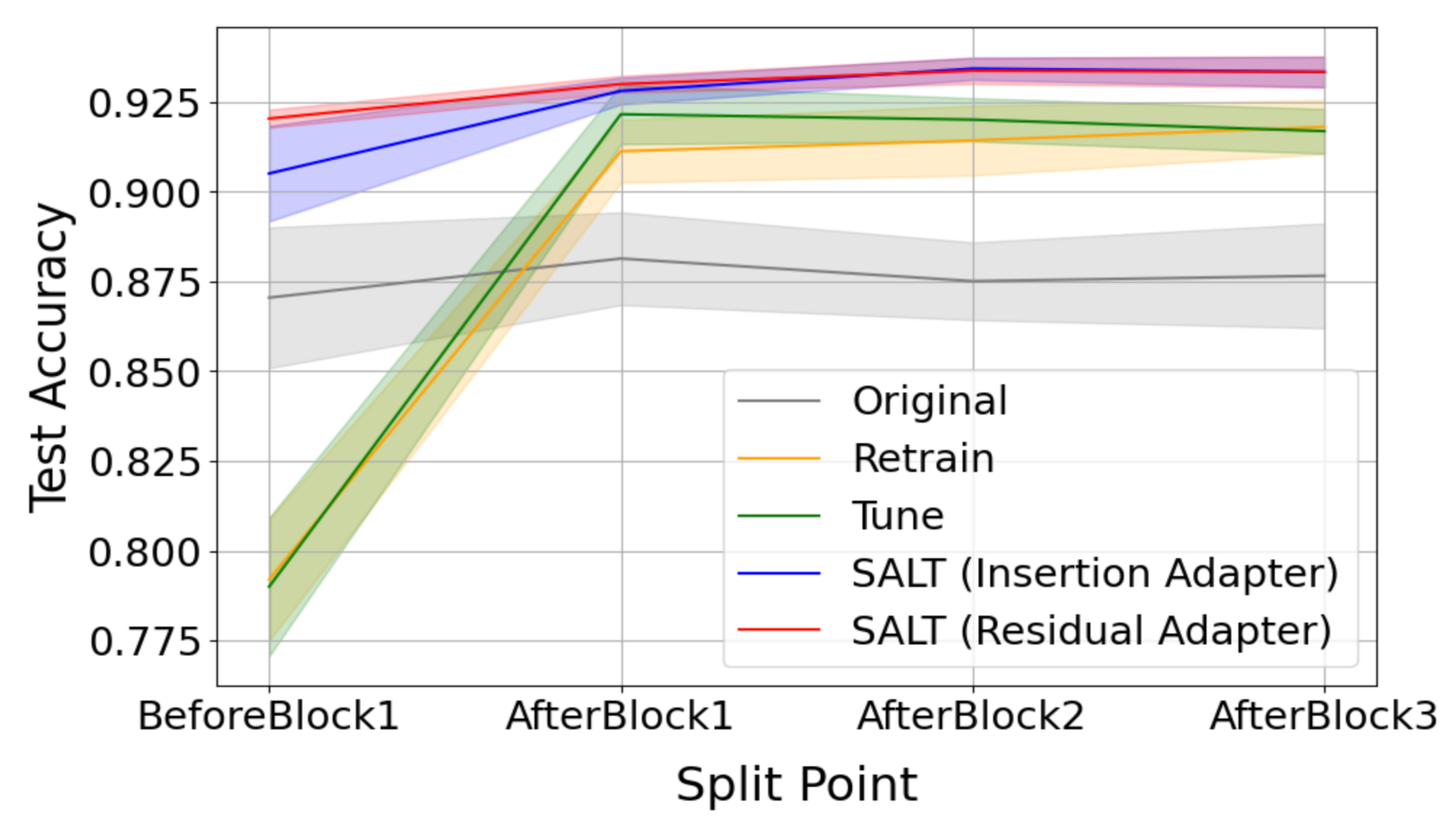}}
    \caption{Accuracy across different split points for each method on CIFAR-10. }
    \label{fig:splitpoint_accuracy}
    \vspace{-2mm}
\end{figure}

Figure~\ref{fig:splitpoint_accuracy} shows personalized accuracy across split points (BeforeBlock1 to AfterBlock3).
SALT consistently outperforms baseline methods regardless of split depth.
For Retrain and Tune, deeper splits increase client-side trainable parameters, enhancing performance through higher model capacity.
In contrast, SALT keeps head and tail networks fixed, and benefits from richer latent features at deeper splits, enabling more effective adaptation.
These results highlight distinct adaptation mechanisms and confirm SALT’s robustness across different abstraction levels.

\section{Conclusion}
\label{sec:Conclusion}
We proposed SALT, a lightweight model adaptation framework for Split Computing under closed constraints, where the pre-trained head and tail networks are proprietary and inaccessible to users. SALT addresses the challenge of adapting models without access to their internal parameters or architectures by introducing a compact, trainable adapter on the client side to refine latent features from the head network, enabling user-specific adaptation without modifying the original models or increasing communication overhead. Experiments on CIFAR-10 and CIFAR-100 demonstrate that SALT achieves improved accuracy with lower training cost compared to fine-tuning methods, while maintaining effectiveness across different split points and showing strong robustness under simulated packet loss. In summary, SALT provides a practical, scalable, and efficient solution for personalized inference in real-world edge AI systems operating under strict system constraints.

\bibliographystyle{IEEEtran}
\bibliography{references}

\end{document}